\documentclass[letterpaper]{article}
\usepackage{amsmath}
\usepackage{amssymb}
\usepackage{aaai2027}

\usepackage{booktabs}
\usepackage{multirow}
\usepackage{graphicx}
\usepackage{caption}

\usepackage[hyphens]{url}
\usepackage{graphicx}
\usepackage{natbib}
\usepackage{caption}
\usepackage{amsmath}
\usepackage{amssymb}
\usepackage{comment}

\usepackage{booktabs}
\usepackage{multirow}

\usepackage{algorithm}
\usepackage{algorithmic}

\usepackage{newfloat}
\usepackage{listings}
\DeclareCaptionStyle{ruled}{labelfont=normalfont,labelsep=colon,strut=off}
\floatstyle{ruled}
\newfloat{listing}{tb}{lst}{}
\floatname{listing}{Listing}

\title{High-accuracy Low-Bit KV-Cache Quantization via Local Distribution Restoration}
\author{
    Gradwell Dzikanyanga$^{1}$†, Yanqi Pan$^{1}$, Weihao Yang$^{1}$, Donglei Wu$^{1}$, Wen Xia$^{1}$‡, Hao Huang$^{1}$
}
\affiliations{
    $^{1}$Harbin Institute of Technology, Shenzhen

    †: \texttt{gdzikanyanga@gmail.com}, ‡: \texttt{xiawen@hit.edu.cn}
}

\usepackage{bibentry}

\begin{document}

\maketitle

\begin{abstract}
Long-context large language model inference relies on the KV cache to avoid redundant attention computation, but incurs high memory and bandwidth overheads. Low-bit KV-cache quantization reduces this cost, yet it severely degrade quality; particularly, one-bit quantization reduces accuracy from 84.2\% to 47.8\% on Llama-3.1-8B under RULER. 

Rather than common beliefs that absolute error of logits, we find that the root cause is \emph{structured local misranking}, where the distribution of logits in top-$K$ region is drifted. We thereby propose \emph{local distribution restoration}, a new technique that detects steps with high local distribution risk from quantized-logit features and restores only the selected top-$K$ candidate distribution before token selection. We implement DGAP to achieve local distribution restoration, with efficient risk detcetors and correctors. Expeirments show that on Llama-3.1-8B, DGAP recovers K1V1 RULER accuracy from 47.8\% to 83.2\% and reduces distribution drift from 0.38 to 0.14; across Llama, Mistral, and Qwen models, it preserves the persistent low-bit KV-cache footprint with modest decode overhead.


\end{abstract}


\section{Introduction}

Large language models (LLMs) increasingly rely on long-context inference for document understanding, retrieval-augmented generation, reasoning, and code analysis~\cite{deepseek-r1,openai}. However, the KV cache grows linearly with context length and batch size; a Llama-style 8B model already requires about 8 GB of full-precision (FP16) KV cache for a 128K-token sequence, making KV-cache memory and bandwidth key serving bottlenecks~\cite{pagedattention,lserve}. KV-cache quantization reduces this cost by storing cached keys and values at low precision, with recent methods improving quantization through asymmetric formats~\cite{kivi}, outlier handling~\cite{kvquant}, and mixed-precision or error-compensated designs~\cite{pmkvq,turboquant}.

Despite these advances, our experiments show that aggressive low-bit KV quantization can still severely degrade quality. Under K1V1, where both keys and values are stored at 1 bit, RULER accuracy on Llama-3.1-8B drops from 84.2\% to 47.8\%, with similar trends on Mistral-7B and Qwen2.5-14B. This motivates a closer analysis of how low-bit KV quantization changes the local candidate distribution before token selection, beyond what memory usage, throughput, and final task scores reveal.

We find that this degradation is not explained by logit-magnitude error alone. In scaled dot-product attention, attention weights are computed as
$\mathrm{softmax}(QK^\top/\sqrt{d_k})V$~\cite{attention, think, AsymKV}.
Therefore, key quantization can directly change query-key scores, reorder
attended cache positions, and alter the context mixture used to form
next-token logits. A controlled analysis that separates score magnitude from candidate ranking shows that imposing the quantized ranking on FP16 scores causes a 31.8-point accuracy drop, whereas ranking-preserving score changes remain within 1.2 points of FP16. These results suggest that aggressive low-bit quantization mainly harms decoding by reordering the local high-probability candidate distribution, rather than by uniformly shifting logit values.

Importantly, this distribution shift is often locally recoverable. Although quantization distorts probability mass and reorders high-probability candidates, those candidates are often not removed from consideration; instead, they remain within a compact quantized top-$K$ candidate region. We call this phenomenon \emph{structured local misranking}. It suggests a practical alternative to increasing persistent KV precision or correcting the full vocabulary: restore only the affected local candidate distribution before token selection.

We propose \textbf{DGAP} (\textbf{D}isagreement-\textbf{G}uided \textbf{A}daptive \textbf{P}recision), a lightweight layer for restoring local distribution fidelity in low-bit KV-cache decoding. DGAP contains two trainable modules. A \textbf{distribution-risk detector} estimates local distribution risk from quantized-logit features, while a \textbf{selective top-$K$ logit corrector} restores the selected candidate distribution only when the risk exceeds a calibrated threshold. This separation keeps runtime overhead low: the detector avoids unnecessary correction on stable steps, and the corrector updates only a compact candidate region rather than the full vocabulary. The remaining logits and the persistent low-bit KV cache are left unchanged.

DGAP is an add-on to existing low-bit KV-cache decoders and requires no base-model retraining or attention-kernel modification. Across Llama-3.1-8B, Mistral-7B, and Qwen2.5-14B on LongBench, RULER, MMLU, and WikiText-2, DGAP improves quality and local distribution fidelity across K1V1--K8V8 settings. Under K1V1 on Llama-3.1-8B, DGAP raises RULER accuracy from 47.8\% to 83.2\%, close to the 84.2\% FP16 result, reduces distribution drift from 0.38 to 0.14, and preserves the persistent 1-bit KV-cache footprint with only ($1.06\times$) relative decode latency.

This work provides a mechanistic analysis of aggressive low-bit KV-cache quantization at the local distribution level. DGAP turns this analysis into a practical restoration mechanism, showing that recoverable candidate-distribution reordering can be detected and corrected locally without reconstructing FP16 KV states or increasing persistent KV precision.

In summary, this paper makes the following contributions:
\begin{itemize}
    \item We provide a mechanistic analysis of aggressive low-bit KV-cache quantization, showing that quality degradation is strongly associated with local candidate-distribution reordering rather than logit-magnitude error alone.
    
    \item We identify \emph{structured local misranking}, where high-probability candidates are reordered by quantization but often remain within a compact quantized top-$K$ region, making local distribution restoration possible.
    
    \item We propose \textbf{DGAP}, a lightweight distribution-restoration layer with a risk detector and selective top-$K$ corrector that restores local candidate distributions without changing the persistent low-bit KV cache.
    
    \item We evaluate DGAP across Llama, Mistral, and Qwen models on long-context and language benchmarks, showing improved quality and distribution fidelity across K1V1--K8V8 with modest decode overhead.

\end{itemize}

\section{Related Work}
\label{sec:related_work}

\paragraph{Long-context KV-cache efficiency.}
Long-context LLM inference is often constrained by KV-cache memory and bandwidth, since cached keys and values grow linearly with sequence length and batch size~\cite{flexgen}. Prior work improves serving efficiency through paged cache management~\cite{pagedattention,lserve}, I/O-aware attention kernels~\cite{flashattention}, and sparse KV retention or eviction based on attention sinks, heavy hitters, or prompt-level importance~\cite{streamingllm,h2o,snapkv,pyramidkv}. These methods reduce memory fragmentation, memory traffic, or the number of retained KV states. However, they mainly optimize how KV states are stored, accessed, or selected, and do not directly analyze how aggressive low-bit KV representation changes the local next-token candidate distribution during decoding.

\paragraph{KV-cache quantization and distribution preservation.}
KV-cache quantization reduces decoding memory and bandwidth by storing cached keys and values at low precision. Existing methods improve low-bit quality through asymmetric key/value quantization~\cite{kivi}, outlier-aware and non-uniform quantization~\cite{kvquant}, sparse or low-rank error compensation~\cite{gear}, progressive mixed precision~\cite{pmkvq}, online vector quantization~\cite{turboquant}, and hierarchical quantized caches for speculative decoding~\cite{quantspec}. A related line of quantization-aware or distribution-aware methods uses calibration, loss-aware objectives, rotations, or distillation to reduce mismatch between low-precision and full-precision models~\cite{llm-qat,leanquant,dartquant}. These approaches primarily improve the quantized representation during quantizer design, calibration, or training, while runtime local distribution drift and candidate reordering under aggressive low-bit KV-cache decoding remain less explored.

\begin{figure}[t]
    \centering
    \includegraphics[width=\columnwidth]{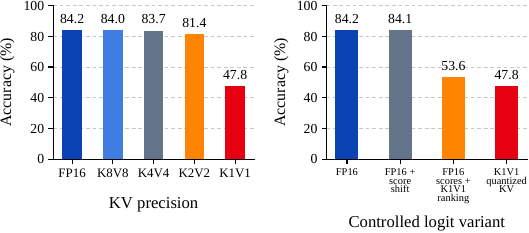}
    \caption{
    \textit{Left}: K1V1 sharply degrades RULER accuracy.
    \textit{Right}: controlled variants identify candidate reordering as the main cause.
}

    \label{fig:motivation_drift_quality}
\end{figure}












\section{Motivation}
\label{sec:motivation}

\subsection{Accuracy Loss under Aggressive KV Quantization}
\label{sec:motivation_low_accuracy}

Although KV-cache quantization reduces memory footprint~\cite{kivi,shadowkv}, aggressive low-bit settings can severely degrade model quality.

\paragraph{Methodology.}
We evaluate Llama-3.1-8B on RULER under FP16 and four KV precision settings: K8V8, K4V4, K2V2, and K1V1, where K$b$V$b$ denotes $b$-bit keys and $b$-bit values. All experiments use NVIDIA A100-80GB GPUs and the setup in Section~\ref{sec:experimental_setup}.

\paragraph{Observation \#1: Aggressive low-bit KV quantization sharply reduces task accuracy.}
As shown in Fig.~\ref{fig:motivation_drift_quality}(left), K1V1 reduces RULER accuracy from 84.2\% under FP16 to 47.8\%. Higher-bit settings remain close to FP16, with K8V8, K4V4, and K2V2 achieving 84.0\%, 83.7\%, and 81.4\%, respectively. Thus, task degradation becomes especially severe in the aggressive 1-bit KV-cache regime.

\begin{figure}[t]
    \centering
    \includegraphics[width=\columnwidth]{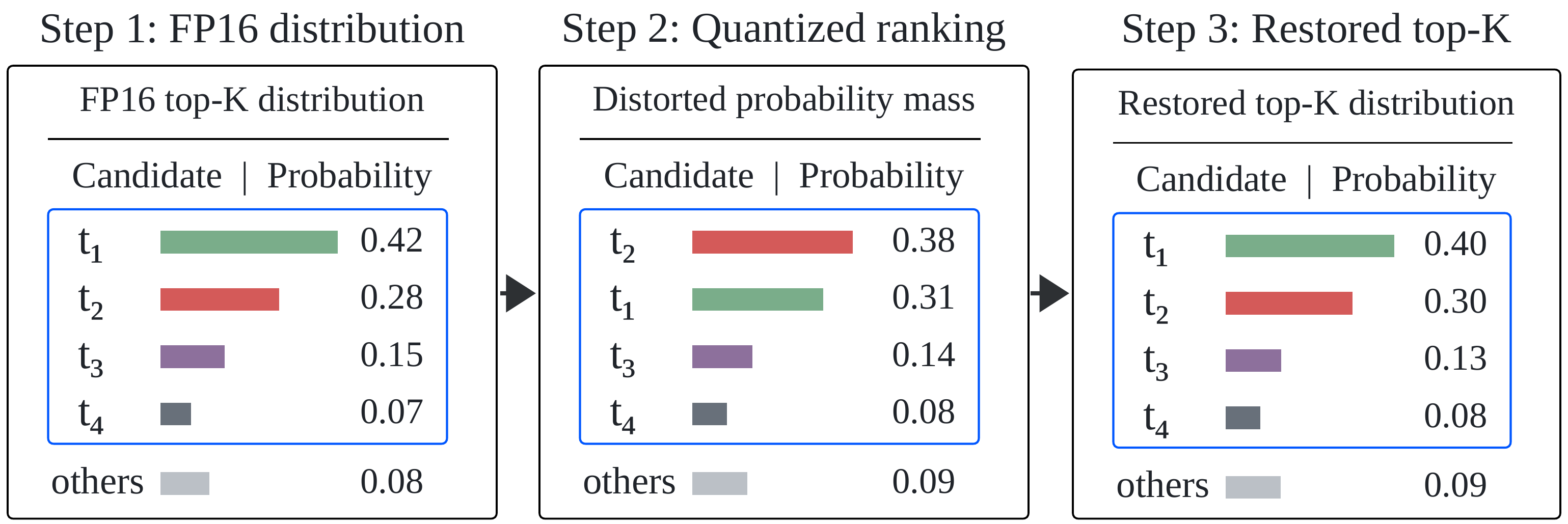}
    \caption{Top-$K$ restoration. Quantization distorts the local top-$K$ distribution, while restoration recovers the selected candidates.
    }
    \label{fig:restoration}
\end{figure}

\subsection{Decode-Time Analysis of Accuracy Loss}
\label{sec:motivation_analysis}

The sharp task drop motivates a local decode-time question: how does low-bit KV quantization alter the high-probability candidate distribution before token selection?

\paragraph{Analysis.}
We consider two possible causes. First, quantization may change logit magnitudes while preserving the relative ordering of high-probability candidates. Second, it may distort probability mass and reorder the local candidate distribution. Because attention weights are computed from query-key scores, key quantization can directly change the ranking of attended cache positions, which in turn alters the context mixture used to form next-token logits. Thus, local candidate reordering can be more damaging than logit-magnitude error alone.

\paragraph{Methodology.}
To separate logit-magnitude changes from candidate-ranking changes, we construct two controlled variants from FP16 logits. The first keeps the FP16 logit values but permutes them according to the quantized ranking, isolating the effect of candidate reordering. The second applies a ranking-preserving score transformation to FP16 logits. In our control, this is implemented as a constant offset, which changes raw logit values but leaves the softmax distribution and token ranking unchanged.

\paragraph{Observation \#2: Local distribution reordering dominates logit-value changes.}
As shown in Fig.~\ref{fig:motivation_drift_quality}, the distribution-preserving variant remains within 1.2 points of FP16, while replacing the FP16 ranking with the quantized ranking causes a 31.8-point accuracy drop. This indicates that quality loss under aggressive low-bit KV quantization is driven mainly by local candidate-distribution reordering, rather than logit-magnitude changes alone. Fig.~\ref{fig:restoration} illustrates this structured local misranking: quantization reorders high-probability FP16 candidates, but many of these candidates remain within a compact quantized top-$K$ region. Therefore, preserving local candidate-distribution fidelity and relative rank structure is crucial for aggressive low-bit KV decoding.

A simple pairwise analysis explains why such reordering changes the local distribution. For two candidates $i$ and $j$ with $z_i^{\mathrm{fp}} > z_j^{\mathrm{fp}}$, the FP16 distribution assigns a larger probability to $i$ than to $j$, since
\[
\frac{p_i^{\mathrm{fp}}}{p_j^{\mathrm{fp}}} = \exp(z_i^{\mathrm{fp}} - z_j^{\mathrm{fp}}) > 1.
\]
Under quantized logits $\hat{z}_i = z_i^{\mathrm{fp}} + e_i$, the corresponding probability ratio becomes
\[
\frac{\hat{p}_i}{\hat{p}_j} =
\exp\big((z_i^{\mathrm{fp}} - z_j^{\mathrm{fp}}) + (e_i - e_j)\big).
\]
The local ordering reverses when
\[
\hat{z}_j > \hat{z}_i
\quad \Leftrightarrow \quad
e_j - e_i > z_i^{\mathrm{fp}} - z_j^{\mathrm{fp}}.
\]
Thus, even moderate non-uniform errors can redistribute probability mass and reorder candidates when the FP16 margin is small. In contrast, adding a constant offset $c$ gives
\[
\operatorname{softmax}(\mathbf{z}^{\mathrm{fp}} + c\mathbf{1}) =
\operatorname{softmax}(\mathbf{z}^{\mathrm{fp}}),
\]
so token probabilities and ranking remain unchanged.

\subsection{Key Idea: Local Distribution Restoration}
\label{sec:motivation_key_idea}

The analysis above suggests that quality loss from aggressive low-bit KV-cache quantization can be mitigated by restoring the affected local top-$K$ candidate distribution before token selection.
A naive distribution-restoration pipeline would involve three steps.
First, compare the quantized decoding distribution with an FP16 reference to identify whether the current step has drifted.
Second, search the full vocabulary to locate tokens whose probability mass or relative ordering has changed. Third, adjust the affected logits so that the resulting distribution better matches the FP16 distribution.

\paragraph{Challenges.}
Although this pipeline is conceptually straightforward, distribution restoration is costly inside autoregressive decoding. First, drift detection cannot rely on FP16 logits or extra reference forward passes at runtime, since this would defeat the purpose of low-bit decoding.
Second, full-vocabulary restoration is impractical because searching and correcting tens of thousands of logits at every decoding step would add substantial compute and memory traffic.
Third, correction must remain local and predictable; falling back to higher persistent KV precision or applying dense vocabulary-wide updates would undermine the memory and latency gains of quantization. These challenges motivate a lightweight detector--corrector framework that identifies risky decoding steps from quantized logits and restores only a compact top-$K$ candidate distribution.

\begin{figure*}[t] 
    \centering
    \includegraphics[width=1\textwidth]{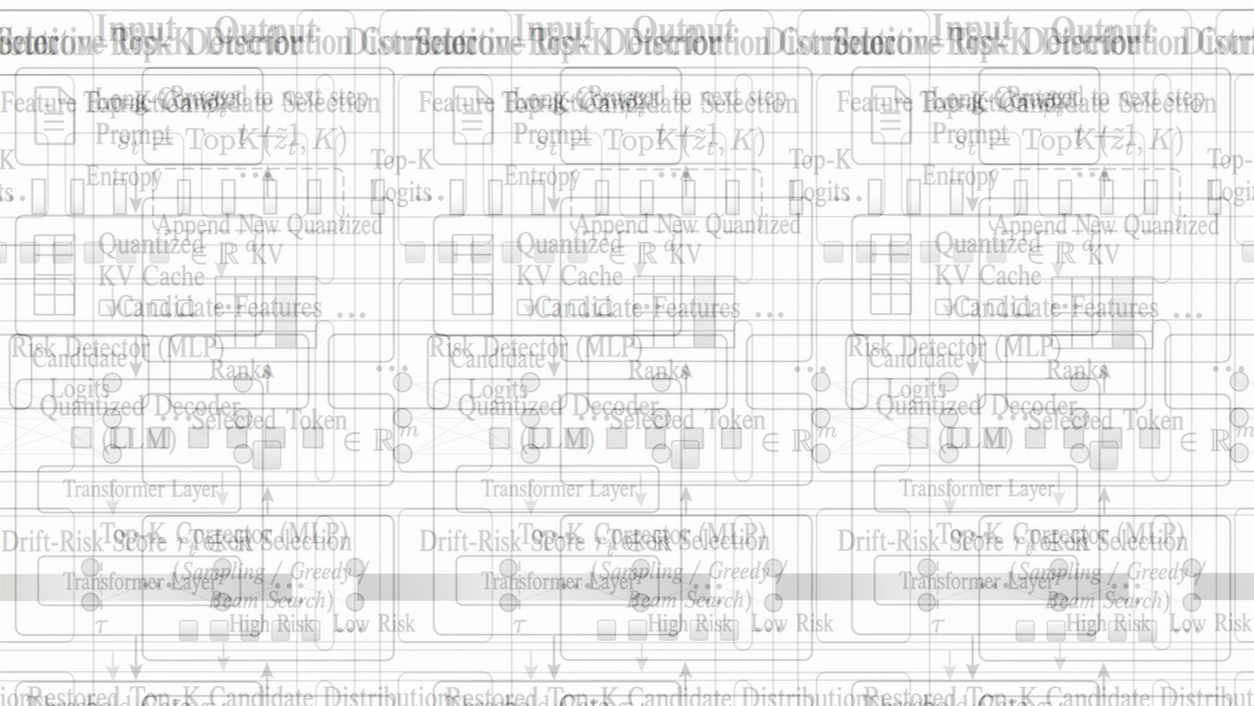}
    \caption{
    DGAP runtime distribution-restoration pipeline.
    At each decoding step, DGAP estimates distribution risk from quantized logits $\hat{\mathbf{z}}_t$ and, only when $r_t>\tau$, restores the selected top-$K$ candidate slice before token selection; otherwise, the original logits are used unchanged.
    }
    \label{fig:dgap_runtime}
\end{figure*}

\section{Methodology}
\label{sec:methodology}

\subsection{DGAP Overview}
\label{sec:dgap_overview}

We propose DGAP (\textbf{D}isagreement-\textbf{G}uided \textbf{A}daptive \textbf{P}recision), a lightweight local distribution-restoration layer for low-bit KV-cache decoding.

\paragraph{Online restoration framework.}
As shown in Fig.~\ref{fig:dgap_runtime}, DGAP operates on logits produced by an existing low-bit KV-cache decoder. At decoding step \(t\), the quantized decoder outputs \(\hat{\mathbf{z}}^{(b)}_t \in \mathbb{R}^{V}\) under precision \(b\). For analysis, we write
\begin{equation}
\hat{\mathbf{z}}^{(b)}_t = \mathbf{z}^{\mathrm{fp}}_t + \boldsymbol{\delta}_t,
\label{eq:quant_error_view}
\end{equation}
where \(\mathbf{z}^{\mathrm{fp}}_t\) denotes the FP16 logits and \(\boldsymbol{\delta}_t\) denotes the quantization-induced logit error. DGAP does not attempt to estimate the full error vector \(\boldsymbol{\delta}_t\). Instead, Eq.~\eqref{eq:quant_error_view} is used only to describe the mismatch between FP16 and low-bit decoding; the goal is to identify and restore the affected local top-\(K\) candidate distribution. During online inference, neither \(\mathbf{z}^{\mathrm{fp}}_t\) nor \(\boldsymbol{\delta}_t\) is available, so DGAP estimates distribution risk using only features derived from \(\hat{\mathbf{z}}^{(b)}_t\).

DGAP has two trainable modules: a \emph{distribution-risk detector} and a \emph{selective top-\(K\) distribution corrector}. Given runtime features \(\boldsymbol{\phi}_t = \Phi(\hat{\mathbf{z}}^{(b)}_t)\), the detector predicts a local distribution-risk score \(r_t\), indicating whether the quantized candidate distribution is likely to be reordered or distorted. If \(r_t \leq \tau\), DGAP uses \(\hat{\mathbf{z}}^{(b)}_t\) directly for token selection. If \(r_t > \tau\), DGAP selects the active local candidate region
\begin{equation}
\mathcal{S}^{(b)}_t = \operatorname{TopK}(\hat{\mathbf{z}}^{(b)}_t, K_b),
\end{equation}
predicts updates only for tokens in \(\mathcal{S}^{(b)}_t\), and merges the corrected slice back into the full logit vector before token selection. Thus, the detector runs at every step, while the corrector is invoked only on high-risk steps to restore the local candidate distribution and relative ordering. Thus, the detector runs at every step, while the corrector is invoked only on high-risk steps to restore the local candidate distribution and relative ordering. The persistent low-bit KV cache is never modified.

\paragraph{Offline supervision.}
\label{sec:dgap_training}
DGAP is trained offline from paired FP16 and quantized-KV decoding traces constructed from the PG-19 training split~\cite{pg-19}. We use PG-19 because it contains naturally long documents, making it suitable for collecting long-context KV-cache traces without using examples from the downstream evaluation benchmarks. The calibration sequences are split into trace-training and validation subsets, both disjoint from LongBench, RULER, MMLU, and WikiText-2 evaluation data. To expose DGAP to different cache lengths, we sample prefixes at multiple context lengths, such as 8K, 16K, 32K, and 64K tokens, when supported by the backbone model.

For each PG-19 sequence, we run the same model twice with the same prefix and target tokens: once with an FP16 KV cache and once with the target low-bit KV cache. At each decoding step \(t\), we record the FP16 logits \(\mathbf{z}^{\mathrm{fp}}_t\), the quantized logits \(\hat{\mathbf{z}}^{(b)}_t\), and the corresponding local candidate distributions. These paired traces provide distribution-risk labels for the detector and local top-\(K\) distribution targets for the corrector. After training, only the detector \(f_{\theta}\), corrector \(g_{\psi}\), calibrated threshold \(\tau\), and recovery-window sizes \(K_b\) are used at runtime; the FP16 teacher is not used during inference.

\paragraph{Adaptive recovery-window selection.}
\label{sec:recovery_window_selection}
The recovery window required to cover the affected local candidate distribution depends primarily on key precision. Lower-bit keys alters query-key attention scores more strongly, increasing local candidate reordering and requiring larger recovery windows. Value precision can affect the aggregated representation, but the dominant source of ranking instability comes from key-induced attention-score changes. For each precision \(b \in \{\mathrm{K1V1},\mathrm{K2V2},\mathrm{K4V4},\mathrm{K8V8}\}\), we calibrate \(K_b\) from paired FP16 and quantized-KV validation traces. Let \(\mathbf{p}^{\mathrm{fp}}_t=\mathrm{softmax}(\mathbf{z}^{\mathrm{fp}}_t)\), and let
\(\pi^{\mathrm{fp}}_t\) denote the FP16 token order sorted by decreasing probability. We define
\(\mathcal{C}^{\mathrm{fp}}_t(\alpha)\) as the smallest prefix of this FP16 order whose cumulative probability mass reaches \(\alpha\):
\begin{equation}
\begin{aligned}
\mathcal{C}^{\mathrm{fp}}_t(\alpha)
&=
\{\pi^{\mathrm{fp}}_t(m)\}_{m=1}^{M_t},\\
M_t
&=
\min\Bigg\{
M:
\sum_{m=1}^{M}
p^{\mathrm{fp}}_{t,\pi^{\mathrm{fp}}_t(m)}
\ge \alpha
\Bigg\}.
\end{aligned}
\end{equation}
We choose the smallest \(K_b\) whose quantized top-\(K\) region covers the target FP16 candidate mass \(\rho_K\):
\begin{equation}
K_b =
\min \left\{
K :
\mathbb{E}_t
\left[
\sum_{i \in \mathcal{C}^{\mathrm{fp}}_t(\alpha)
\cap
\operatorname{TopK}(\hat{\mathbf{z}}^{(b)}_t,K)}
p^{\mathrm{fp}}_{t,i}
\right]
\ge \rho_K
\right\}.
\end{equation}
At runtime, the active precision selects \(K_b\), and DGAP corrects only
\(\mathcal{S}^{(b)}_t=\operatorname{TopK}(\hat{\mathbf{z}}^{(b)}_t,K_b)\).
Lower-bit settings therefore use larger recovery windows, while higher-bit settings use smaller ones, avoiding both full-vocabulary correction and uniformly higher persistent KV precision.

\paragraph{Why local candidate reordering occurs.}
The compact recovery region follows from Eq.~\ref{eq:quant_error_view}. Since attention weights are computed from query-key scores, key quantization can reorder attended cache positions and induce non-uniform logit errors. For candidates \(i,j\) with \(z^{\mathrm{fp}}_{t,i}>z^{\mathrm{fp}}_{t,j}\), define \(m_{t,ij}=z^{\mathrm{fp}}_{t,i}-z^{\mathrm{fp}}_{t,j}\). Their order reverses when
\begin{equation}
\delta_{t,j}-\delta_{t,i}>m_{t,ij}.
\end{equation}
Thus, small-margin high-probability candidates are most vulnerable to reordering. This matters because decoding depends on relative probabilities rather than absolute logit error: an order-preserving shift may leave token selection nearly unchanged, whereas a non-uniform error that reverses local candidates can move probability mass to a different mode and alter subsequent autoregressive states. This motivates local top-\(K\) restoration rather than full-vocabulary correction.

\subsection{Distribution-Risk Detector}
\label{sec:risk_detector}

\paragraph{Goal.}
During online inference, the FP16 logits \(\mathbf{z}^{\mathrm{fp}}_t\) and quantization-induced error \(\boldsymbol{\delta}_t\) are unavailable. The detector therefore predicts whether the current low-bit decoding step is likely to exhibit recoverable local candidate-distribution drift using only features from the quantized logits. Given lightweight features \(\boldsymbol{\phi}_t=\Phi(\hat{\mathbf{z}}^{(b)}_t)\), it predicts
\begin{equation}
r_t=\sigma\!\left(f_{\theta}(\boldsymbol{\phi}_t)\right),
\qquad r_t\in[0,1],
\end{equation}
where \(f_{\theta}\) is the detector and \(\sigma(\cdot)\) is the sigmoid function. Correction is triggered only when \(r_t>\tau\), with \(\tau\) calibrated offline on validation traces.

\paragraph{Runtime features.}
The detector uses compact distribution-shape statistics from \(\hat{\mathbf{z}}^{(b)}_t\), including entropy, top-\(K_b\) probability mass, local gap statistics, probability-mass concentration, and dispersion among high-probability candidates. These features capture uncertainty and local rank instability: small candidate gaps or dispersed probability mass make non-uniform quantization errors more likely to reorder the local distribution. All features are computed from quantized logits, without FP16 logits or extra forward passes.

\paragraph{Offline supervision and calibration.}
We train the detector from paired FP16 and quantized-KV traces. Let
\(\mathbf{p}^{\mathrm{fp}}_t=\operatorname{softmax}(\mathbf{z}^{\mathrm{fp}}_t)\) and
\(\hat{\mathbf{p}}^{(b)}_t=\operatorname{softmax}(\hat{\mathbf{z}}^{(b)}_t)\) denote the FP16 and quantized next-token distributions. Let \(\mathcal{C}^{\mathrm{fp}}_t(\alpha)\) be the smallest FP16 candidate set whose cumulative probability mass reaches \(\alpha\), and let \(\mathcal{S}^{(b)}_t=\operatorname{TopK}(\hat{\mathbf{z}}^{(b)}_t,K_b)\) be the active quantized recovery region. We measure recoverable candidate coverage as
\begin{equation}
c_t =
\sum_{i \in \mathcal{C}^{\mathrm{fp}}_t(\alpha)\cap \mathcal{S}^{(b)}_t}
p^{\mathrm{fp}}_{t,i}.
\end{equation}
To measure local distribution distortion, we compare normalized probability vectors over
\(\mathcal{U}_t=\mathcal{C}^{\mathrm{fp}}_t(\alpha)\cup \mathcal{S}^{(b)}_t\):
\begin{equation}
d^{\mathrm{loc}}_t =
\left\|
\bar{\mathbf{p}}^{\mathrm{fp}}_{t,\mathcal{U}_t}
-
\bar{\mathbf{p}}^{(b)}_{t,\mathcal{U}_t}
\right\|_2 ,
\end{equation}
where \(\bar{\mathbf{p}}_{\mathcal{U}}\) denotes the probability vector renormalized over \(\mathcal{U}\). A step is labeled risky when its local candidate distribution is distorted but sufficient FP16 candidate mass remains recoverable:
\begin{equation}
y_t =
\mathbb{I}
\left[
d^{\mathrm{loc}}_t>\epsilon
\;\land\;
c_t\ge\rho_c
\right].
\label{eq:recoverable_risk_label}
\end{equation}
Distorted steps with low candidate coverage are not labeled positive because the FP16-relevant candidates are largely absent from the quantized recovery window, making sparse local correction unreliable. Here, \(\rho_K\) is used to choose \(K_b\), while \(\rho_c\) determines whether a distorted step is recoverable enough for detector supervision. The detector is trained with binary cross-entropy:
\begin{equation}
\mathcal{L}_{\mathrm{risk}} =
-\sum_t
\left[
y_t\log r_t+(1-y_t)\log(1-r_t)
\right].
\label{eq:risk_loss}
\end{equation}
After training, \(\tau\) is calibrated offline to trade off restoration quality and correction frequency.

\subsection{Selective Top-$K$ Distribution Corrector}
\label{sec:topk_corrector}

\paragraph{Goal.}
For high-risk steps, the corrector restores the local candidate distribution before token selection. Correcting all logits \(\hat{\mathbf{z}}^{(b)}_t \in \mathbb{R}^{V}\) would require full-vocabulary computation and reduce the efficiency gains of low-bit KV-cache decoding. Motivated by the local reordering analysis in Sec.~\ref{sec:motivation}, DGAP corrects only the quantized top-\(K_b\) recovery window, which is calibrated to cover the high-probability FP16 candidate mass.

\paragraph{Local candidate correction.}
When \(r_t > \tau\), DGAP selects the active recovery window
\begin{equation}
\mathcal{S}^{(b)}_t = \operatorname{TopK}(\hat{\mathbf{z}}^{(b)}_t, K_b),
\qquad K_b \ll V .
\end{equation}
For tokens in \(\mathcal{S}^{(b)}_t\), we construct candidate-level features
\(\boldsymbol{\xi}_{t,\mathcal{S}^{(b)}_t}\) from local logits, ranks, margins, and probabilities. The corrector \(g_{\psi}\) predicts sparse updates for this slice:
\begin{equation}
\Delta\hat{\mathbf{z}}_{t,\mathcal{S}^{(b)}_t} =
g_{\psi}(\boldsymbol{\xi}_{t,\mathcal{S}^{(b)}_t}, \boldsymbol{\phi}_t),
\end{equation}
where \(\boldsymbol{\phi}_t\) denotes the detector features. The corrected logits are obtained by sparse merge:
\begin{equation}
\tilde{z}_{t,j} =
\begin{cases}
\hat{z}^{(b)}_{t,j} + \Delta\hat{z}_{t,j}, & j \in \mathcal{S}^{(b)}_t \ \text{and}\ r_t > \tau, \\
\hat{z}^{(b)}_{t,j}, & \text{otherwise}.
\end{cases}
\label{eq:delta_merge}
\end{equation}
Thus, non-selected logits remain unchanged; if \(r_t \leq \tau\), DGAP bypasses the corrector and uses \(\tilde{\mathbf{z}}_t=\hat{\mathbf{z}}^{(b)}_t\).

\paragraph{Training objective.}
The corrector is trained offline using the same paired FP16 and quantized-KV traces as the detector. For compactness, let \(S=\mathcal{S}^{(b)}_t\). For each selected window, we form:
\begin{equation}
\begin{aligned}
\mathbf{q}^{\mathrm{fp}}_{t,S}
&=\mathrm{softmax}(\mathbf{z}^{\mathrm{fp}}_{t,S}/T),\quad
\tilde{\mathbf{q}}_{t,S}
=\mathrm{softmax}(\tilde{\mathbf{z}}_{t,S}/T),
\end{aligned}
\end{equation}
where \(T\) is a training temperature. We match the corrected local distribution to the FP16 local distribution:
\begin{equation}
\mathcal{L}_{\mathrm{dist}} =
\sum_t
y_t
D_{\mathrm{KL}}
\left(
\mathbf{q}^{\mathrm{fp}}_{t,S}
\middle\|
\tilde{\mathbf{q}}_{t,S}
\right).
\end{equation}

To directly target structured local reordering, we add a pairwise rank-consistency loss over candidates in the selected window. Let
\begin{equation}
\mathcal{P}^{\mathrm{fp}}_t
=
\left\{
(i,j): i,j\in\mathcal{S}^{(b)}_t,\ 
z^{\mathrm{fp}}_{t,i}>z^{\mathrm{fp}}_{t,j}
\right\}
\end{equation}
be the set of FP16-ordered candidate pairs. We define
\begin{equation}
\mathcal{L}_{\mathrm{rank}}
=
\sum_t y_t
\sum_{(i,j)\in\mathcal{P}^{\mathrm{fp}}_t}
w_{t,ij}
\log
\left(
1+
\exp
\left(
-(\tilde{z}_{t,i}-\tilde{z}_{t,j})
\right)
\right),
\end{equation}
where \(w_{t,ij}=p^{\mathrm{fp}}_{t,i}-p^{\mathrm{fp}}_{t,j}\) weights pairwise ordering errors by their FP16 probability gap. This pairwise term is used only during offline training and is restricted to \(K_b\ll V\) candidates, so it does not affect decode-time latency. The final objective is
\begin{equation}
\mathcal{L}_{\mathrm{corr}} =
\mathcal{L}_{\mathrm{dist}}
+
\lambda_{\mathrm{rank}} \mathcal{L}_{\mathrm{rank}}
+
\lambda_{\mathrm{reg}}
\sum_t
y_t
\,
\left\|
\Delta\hat{\mathbf{z}}_{t,\mathcal{S}^{(b)}_t}
\right\|_2^2 .
\label{eq:corr_loss}
\end{equation}
Thus, the KL term restores the local probability shape, the pairwise rank-consistency term preserves relative ordering among high-probability candidates, and the regularizer keeps sparse updates stable.

\paragraph{Why detector--corrector instead of end-to-end correction?}
A single end-to-end corrector could directly map quantized logits to a restored distribution, but it would have to run at every decoding step, including stable steps where correction is unnecessary. If such a model were made conditional to avoid correcting every token, it would require an internal gating mechanism, effectively reintroducing the detector. This would increase compute and memory traffic and weaken the efficiency benefit of low-bit KV-cache decoding. DGAP separates the cheap decision from the heavier restoration: the detector runs every step using compact distribution-shape features, while the corrector is invoked only when the local candidate distribution is distorted but recoverable. The threshold \(\tau\) therefore controls correction frequency, keeping runtime overhead low while preserving the persistent low-bit KV-cache footprint.


\begin{table*}[t]
\centering
\small
\setlength{\tabcolsep}{1.8pt}
\caption{
LongBench results across three instruction-tuned models. DGAP-1bit substantially recovers the degradation of K1V1 while retaining the 1-bit KV-cache setting, surpassing KIVI-2bit and TurboQuant-2.5bit on average and approaching KVQuant-3bit.
}

\label{tab:longbench_three_models}
\begin{tabular}{llccccccccc}
\toprule
\textbf{Model} & \textbf{Method} 
& \textbf{Qasp.} 
& \textbf{HotQA} 
& \textbf{MNews} 
& \textbf{TREC} 
& \textbf{TrivQA} 
& \textbf{SAM} 
& \textbf{GovRpt} 
& \textbf{RepoP} 
& \textbf{Avg} \\
\midrule
\multirow{6}{*}{Llama3.1-8B}
& FP16 \textit{(full-precision ref.)} 
& 9.80 & 13.00 & 18.60 & 69.00 & 88.40 & 43.20 & 28.90 & 59.80 & 41.34 \\
& KIVI-2bit \textit{(2-bit ref. target)} 
& 9.40 & 12.60 & 17.40 & 68.50 & 87.70 & 42.90 & 27.80 & 58.90 & 40.65 \\
& TurboQuant-2.5bit
& 9.45 & 12.65 & 17.65 & 68.70 & 87.85 & 42.95 & 27.90 & 59.05 & 40.78 \\
& KVQuant-3bit 
& 9.60 & 12.80 & 18.00 & 69.00 & 88.10 & 43.10 & 28.20 & 59.40 & 41.02 \\
& K1V1-1bit 
& 3.10 & 8.30  & 7.20  & 54.00 & 73.50 & 33.40 & 14.60 & 44.80 & 29.99 \\
& \textbf{DGAP-1bit (ours)} 
& \textbf{9.50} & \textbf{12.70} & \textbf{17.80} & \textbf{68.80} & \textbf{87.90} & \textbf{43.00} & \textbf{28.00} & \textbf{59.10} & \textbf{40.85} \\
\midrule
\multirow{6}{*}{Mistral-7B}
& FP16 \textit{(full-precision ref.)} 
& 8.10 & 12.70 & 20.00 & 67.50 & 89.80 & 41.70 & 27.60 & 59.00 & 40.80 \\
& KIVI-2bit \textit{(2-bit ref. target)} 
& 7.90 & 12.50 & 18.90 & 66.50 & 89.60 & 41.70 & 27.10 & 58.60 & 40.35 \\
& TurboQuant-2.5bit
& 7.95 & 12.55 & 19.10 & 66.80 & 89.65 & 41.75 & 27.25 & 58.65 & 40.46 \\
& KVQuant-3bit 
& 8.00 & 12.60 & 19.40 & 67.00 & 89.70 & 41.80 & 27.40 & 58.80 & 40.59 \\
& K1V1-1bit 
& 3.60 & 7.90  & 6.80  & 53.00 & 74.80 & 32.80 & 13.90 & 43.70 & 29.56 \\
& \textbf{DGAP-1bit (ours)} 
& \textbf{8.00} & \textbf{12.60} & \textbf{19.30} & \textbf{67.00} & \textbf{89.70} & \textbf{41.80} & \textbf{27.40} & \textbf{58.80} & \textbf{40.58} \\
\midrule
\multirow{6}{*}{Qwen2.5-14B}
& FP16 \textit{(full-precision ref.)} 
& 14.60 & 15.80 & 24.70 & 70.50 & 90.20 & 44.70 & 31.60 & 62.40 & 44.31 \\
& KIVI-2bit \textit{(2-bit ref. target)} 
& 13.90 & 15.10 & 23.80 & 69.80 & 89.60 & 44.10 & 30.40 & 61.20 & 43.49 \\
& TurboQuant-2.5bit
& 14.00 & 15.20 & 23.95 & 69.90 & 89.75 & 44.20 & 30.55 & 61.35 & 43.61 \\
& KVQuant-3bit 
& 14.20 & 15.40 & 24.20 & 70.00 & 89.90 & 44.30 & 30.90 & 61.70 & 43.83 \\
& K1V1-1bit 
& 4.70  & 9.40  & 9.10  & 56.50 & 76.20 & 34.90 & 16.10 & 47.60 & 32.00 \\
& \textbf{DGAP-1bit (ours)} 
& \textbf{14.10} & \textbf{15.30} & \textbf{24.10} & \textbf{70.00} & \textbf{89.90} & \textbf{44.30} & \textbf{30.70} & \textbf{61.50} & \textbf{43.74} \\
\bottomrule
\end{tabular}
\end{table*}

\begin{table}[t]
\centering
\small
\setlength{\tabcolsep}{0.8pt}
\caption{ Generalization beyond LongBench. DGAP-1bit recovers K1V1 degradation across perplexity, knowledge, and controlled long-context benchmarks. }
\label{tab:wikitext_mmlu_ruler}
\begin{tabular}{l l c c c}
\toprule
\textbf{Model} & \textbf{Method} 
& \shortstack{\textbf{WikiText-2}\\ \textbf{PPL}}
& \shortstack{\textbf{MMLU}\\ \textbf{Acc.(\%)}}
& \shortstack{\textbf{RULER}\\ \textbf{Acc.(\%)}} \\
\midrule
\multirow{6}{*}{Llama3.1-8B}
& \textit{FP16 ( ref.)}    
& 8.0  & 78.0 & 84.2 \\
& \textit{KIVI-2bit ( ref. target)}    
& 9.2  & 77.2 & 83.1 \\
& TurboQuant-2.5bit
& 8.9  & 77.4 & 83.3 \\
& KVQuant-3bit 
& 8.8  & 77.5 & 83.5 \\
& K1V1-1bit    
& 37.6 & 42.4 & 47.8 \\
& \textbf{DGAP-1bit}    
& \textbf{9.0}  & \textbf{77.3} & \textbf{83.2} \\
\midrule
\multirow{6}{*}{Mistral-7B}
& \textit{FP16 (ref.)}    
& 12.6 & 75.4 & 82.1 \\
& \textit{KIVI-2bit (ref. target)}    
& 13.8 & 74.6 & 80.9 \\
& TurboQuant-2.5bit
& 13.5 & 74.8 & 81.1 \\
& KVQuant-3bit 
& 13.3 & 74.9 & 81.3 \\
& K1V1-1bit    
& 41.1 & 36.2 & 42.5 \\
& \textbf{DGAP-1bit}    
& \textbf{13.6} & \textbf{74.7} & \textbf{81.0} \\
\midrule
\multirow{6}{*}{Qwen2.5-14B}
& \textit{FP16 (ref.)}    
& 10.9 & 81.2 & 87.1 \\
& \textit{KIVI-2bit (ref. target)}   
& 11.9 & 80.3 & 85.8 \\
& TurboQuant-2.5bit
& 11.6 & 80.5 & 86.1 \\
& KVQuant-3bit 
& 11.5 & 80.6 & 86.2 \\
& K1V1-1bit    
& 32.3 & 51.4 & 56.1 \\
& \textbf{DGAP-1bit}    
& \textbf{11.8} & \textbf{80.4} & \textbf{86.0} \\
\bottomrule
\end{tabular}
\end{table}


\section{Experimental Setup}
\label{sec:experimental_setup}

\paragraph{Testbed.}
All experiments run on NVIDIA A100-80GB GPUs using a unified PyTorch--CUDA implementation. Unless stated otherwise, we use PyTorch SDPA and evaluate batch-size-1 decoding, following the long-context setting where KV-cache memory and decode latency are most critical.

\paragraph{Implementation details.}
We use Python 3.10, PyTorch 2.5.1+cu121, Transformers 4.45.1, CUDA 12.1, and PyTorch SDPA. All methods use the same evaluation prompts, decoding settings, and random seed for each model--benchmark pair. DGAP is trained and calibrated offline on PG-19 traces, while all reported test results are measured on disjoint downstream benchmarks. DGAP hyperparameters, including \(\tau\), \(K_b\), temperature \(T\), and loss weights, are selected on validation traces and fixed for test evaluation. Latency is measured with CUDA synchronization, and we report p95 per-token decode latency and throughput.

\paragraph{Models and benchmarks.}
We evaluate three instruction-tuned decoder-only models: Llama-3.1-8B-Instruct~\cite{llama2}, Mistral-7B-Instruct~\cite{mistral7b}, and Qwen2.5-14B-Instruct~\cite{qwen}. Benchmarks include WikiText-2~\cite{wikitext} for perplexity, MMLU~\cite{mmlu} for knowledge and reasoning, LongBench~\cite{longbench} for practical long-context tasks, and RULER~\cite{ruler} for controlled long-context evaluation.

\paragraph{Baselines.}
We compare DGAP with FP16, uniform low-bit KV baselines, and recent KV-cache quantization methods. FP16 is the full-precision reference. K1V1 is the storage-equivalent aggressive low-bit baseline, while K2V2, K4V4, and K8V8 show the effect of increasing persistent KV precision. We further compare against KIVI-2bit~\cite{kivi}, KVQuant-3bit~\cite{kvquant}, and TurboQuant-2.5bit~\cite{turboquant}, with KIVI-2bit serving as the main strong low-bit reference.

\paragraph{Metrics.}
We report task score, perplexity, KV memory, compression ratio, fallback rate, p95 decode latency, and throughput. To measure local distribution fidelity, we report local distribution drift from FP16. These metrics directly measure whether low-bit quantization reorders the local candidate distribution and whether DGAP restores it.

\begin{table}[t]
\centering
\small
\setlength{\tabcolsep}{2.4pt}
\caption{
DGAP gains are largest under aggressive KV quantization and diminish as bitwidth increases. Scores are averaged over NIAH-2, NIAH-3, MK-2, MK-3, and MV.
}
\label{tab:dgap_bitwidth_ruler_avg}
\begin{tabular}{llccc}
\toprule
\textbf{Model} & \textbf{Setting} & \textbf{Base} & \textbf{+DGAP} & \textbf{Gain} \\
\midrule
\multirow{4}{*}{Llama3.1-8B}
& K1V1 & 53.1 & 85.0 & +31.9 \\
& K2V2 & 85.4 & 87.9 & +2.5 \\
& K4V4 & 88.4 & 89.0 & +0.5 \\
& K8V8 & 88.9 & 89.2 & +0.2 \\
\midrule
\multirow{4}{*}{Mistral-7B}
& K1V1 & 48.1 & 80.7 & +32.6 \\
& K2V2 & 82.7 & 84.9 & +2.3 \\
& K4V4 & 85.0 & 85.5 & +0.6 \\
& K8V8 & 85.6 & 85.8 & +0.2 \\
\midrule
\multirow{4}{*}{Qwen2.5-14B}
& K1V1 & 60.3 & 88.3 & +27.9 \\
& K2V2 & 89.3 & 91.0 & +1.7 \\
& K4V4 & 91.1 & 91.5 & +0.5 \\
& K8V8 & 91.6 & 91.8 & +0.2 \\
\bottomrule
\end{tabular}
\end{table}

\begin{table}[t]
\centering
\small
\setlength{\tabcolsep}{5pt}
\caption{
DGAP improves local distribution fidelity under KV quantization, leading to consistent RULER accuracy gains across bitwidths.
}
\label{tab:distribution_fidelity}
\begin{tabular}{l c c}
\toprule
\textbf{Bit Setting} &
\textbf{Local Distribution Drift} $\downarrow$ &
\textbf{Accuracy} $\uparrow$ \\
\midrule
K1V1 $\rightarrow$ +\textbf{DGAP}
& 0.38 $\rightarrow$ \textbf{0.14}
& 47.8 $\rightarrow$ \textbf{83.2} \\

K2V2 $\rightarrow$ +\textbf{DGAP}
& 0.17 $\rightarrow$ \textbf{0.10}
& 81.4 $\rightarrow$ \textbf{83.6} \\

K4V4 $\rightarrow$ +\textbf{DGAP}
& 0.11 $\rightarrow$ \textbf{0.07}
& 83.7 $\rightarrow$ \textbf{84.0} \\

K8V8 $\rightarrow$ +\textbf{DGAP}
& 0.06 $\rightarrow$ \textbf{0.04}
& 84.0 $\rightarrow$ \textbf{84.2} \\
\midrule
FP16 reference
& 0.00
& 84.2 \\
\bottomrule
\end{tabular}
\end{table}

\subsection{End-task Quality Restoration}
\label{sec:end_task_quality}

We first evaluate whether DGAP restores task quality under aggressive low-bit KV-cache quantization. Tables~\ref{tab:longbench_three_models} and~\ref{tab:wikitext_mmlu_ruler} report results on LongBench, WikiText-2, MMLU, and RULER. On LongBench, \textbf{DGAP-1bit} improves over K1V1 by 36.2\%, 37.3\%, and 36.7\% on Llama-3.1-8B, Mistral-7B, and Qwen2.5-14B, respectively, and exceeds the KIVI-2bit reference on average. On WikiText-2, MMLU, and RULER, DGAP consistently narrows the gap between K1V1 and FP16, reaching higher-bit quality across models. These results show that local distribution restoration recovers much of the quality lost under low-bit KV quantization without increasing persistent KV precision.

\subsection{DGAP Gains Decrease with Increasing KV Precision}
Table~\ref{tab:dgap_bitwidth_ruler_avg} reports DGAP performance across KV bitwidths using averaged scores over NIAH-2, NIAH-3, MK-2, MK-3, and MV. Across all models, DGAP yields its largest improvements under aggressive quantization (K1V1), where performance increases by 27.9--32.6 points over the base setting. As KV precision increases, the gains decrease substantially, with only marginal improvements under K4V4 and K8V8.
This trend indicates that DGAP is most effective in correcting severe low-bit distortion, where local candidate reordering is strongest, while having limited impact when higher-bit KV caches already preserve much of the original decoding behavior.

\begin{table}[t]
\centering
\small
\setlength{\tabcolsep}{5pt}
\caption{
Persistent KV-cache memory (GB) across context lengths for Qwen2.5-14B. DGAP-1bit does not modify KV storage and maintains the same 1-bit footprint as K1V1-1bit, achieving 14.2$\times$ compression over FP16.
}
\label{tab:persistent_kv_memory_qwen}
\begin{tabular}{l c c c c c c}
\toprule
\textbf{Method}
& \shortstack{\textbf{8K}} & \shortstack{\textbf{16K}} & \shortstack{\textbf{32K}} & \shortstack{\textbf{64K}} & \shortstack{\textbf{128K}} & \textbf{Comp.} \\
\midrule

FP16 (ref.)
& 6.73 & 13.46 & 26.92 & 53.85 & 107.69 & 1.0$\times$ \\

KIVI-2bit
& 0.91 & 1.82  & 3.64  & 7.28  & 14.55  & 7.4$\times$ \\

TurboQuant-2.5bit
& 1.05 & 2.10  & 4.21  & 8.41  & 16.83  & 6.4$\times$ \\

KVQuant-3bit
& 1.40 & 2.80  & 5.61  & 11.22 & 22.44  & 4.8$\times$ \\

K1V1-1bit
& 0.47 & 0.95  & 1.90  & 3.79  & 7.58   & 14.2$\times$ \\

\textbf{DGAP-1bit (ours)}
& \textbf{0.47} & \textbf{0.95} & \textbf{1.90} & \textbf{3.79} & \textbf{7.58} & \textbf{14.2$\times$} \\

\bottomrule
\end{tabular}
\end{table}

\subsection{Decode-Time Local Distribution Fidelity}
\label{sec:distribution_fidelity}
We analyze whether DGAP improves decode-time local distribution fidelity. As shown in Table~\ref{tab:distribution_fidelity}, DGAP consistently reduces local distribution drift under KV quantization, leading to improved RULER accuracy.
Improvements become smaller at higher bitwidths because the base KV representations already preserve the FP16 local candidate structure. These results indicate that DGAP restores local distribution fidelity by correcting quantization-induced candidate reordering.

\subsection{Efficiency Analysis}
\label{sec:efficiency}

\paragraph{KV-Cache Memory Across Context Lengths.}
Table~\ref{tab:persistent_kv_memory_qwen} reports KV-cache memory (GB) across context lengths for Qwen2.5-14B. DGAP-1bit maintains the same persistent KV footprint as K1V1-1bit across all context lengths, since it does not modify the stored KV representation. As a result, it preserves a constant 14.2$\times$ compression over FP16.

\begin{figure}[t]
    \centering
    \includegraphics[width=0.85\linewidth,height=0.38\textheight,keepaspectratio]{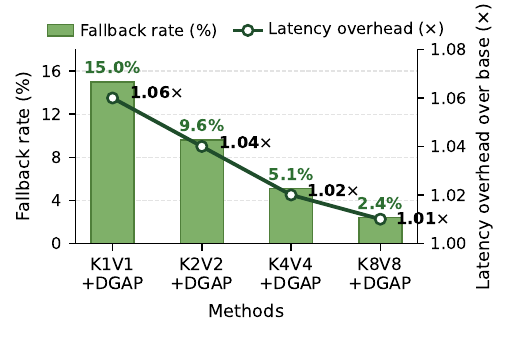}
    \caption{
    DGAP runtime overhead across KV precisions. Lower-bit settings trigger more corrections, while higher-bit settings require fewer fallbacks and incur lower latency overhead. 
    }
    \label{fig:latency_overhead}
\end{figure}

\paragraph{Decode latency and throughput.}
Table~\ref{tab:latency_context} reports batch-size-1 per-token decoding latency under K1V1. DGAP introduces a small overhead of approximately 1.06$\times$, consistent with Fig.~\ref{fig:latency_overhead}. The corresponding throughput drop remains below 6\%, confirming that DGAP preserves efficient decoding.

\paragraph{Why overhead is small.}
The runtime cost of DGAP is limited to lightweight risk detection and selective top-$K$ correction. Since correction is applied only at distribution-risky decoding steps and restricted to local candidate sets, the additional computation remains sparse and does not scale with sequence length.

\begin{table}[t]
\centering
\small
\setlength{\tabcolsep}{3.5pt}
\caption{
Per-token decode latency of DGAP under K1V1 using Qwen2.5-14B.
}
\label{tab:latency_context}
\begin{tabular}{lcccc}
\toprule
\textbf{Context length} & \textbf{8K} & \textbf{16K} & \textbf{32K} & \textbf{64K} \\
\midrule
w/o DGAP (ms)        & 3.20 & 4.85 & 7.40 & 11.80 \\
w/ DGAP (ms)         & 3.40 & 5.14 & 7.84 & 12.46 \\
Extra overhead       & 6.25\% & 5.98\% & 5.95\% & 5.59\% \\
Throughput drop      & 5.88\% & 5.64\% & 5.62\% & 5.29\% \\
\bottomrule
\end{tabular}
\end{table}

\begin{table}[t]
\centering
\small
\setlength{\tabcolsep}{4pt}
\caption{
Ablation on restoration target. DGAP-1bit restores local top-$K$ distribution and surpasses higher-bit KV precision while retaining K1V1 memory footprint.
}
\label{tab:ablation_target}
\begin{tabular}{l c}
\toprule
\textbf{Setting} & \textbf{Avg. Score} \\
\midrule
K1V1        & 53.1 \\
K2V1        & 66.0 \\
K1V2        & 67.3 \\
K2V2        & 81.4 \\
K4V4        & 83.7 \\
K8V8        & 84.0 \\
\midrule
\textbf{DGAP-1bit (ours)} & \textbf{83.2} \\
\bottomrule
\end{tabular}
\end{table}

\subsection{Ablation Studies}
\label{sec:ablation}

We analyze four key design choices in DGAP: (i) the restoration target, (ii) detector--corrector decomposition, and (iii) the risk threshold $\tau$ controlling adaptive correction.

\paragraph{Restoration target: local distribution vs. KV precision.}
We first examine what should be restored under low-bit KV quantization. Unlike methods that increase persistent KV precision, DGAP directly restores the local top-$K$ candidate distribution at decode time. As shown in Table~\ref{tab:ablation_target}, increasing KV precision (K2V2, K4V4) improves performance by reducing information loss, but DGAP-1bit surpasses higher-bit settings while retaining the K1V1 memory footprint. This indicates that correcting local candidate reordering is more effective than uniformly increasing KV precision.

\begin{figure}[H]
    \centering
    \includegraphics[width=0.85\linewidth,height=0.38\textheight,keepaspectratio]{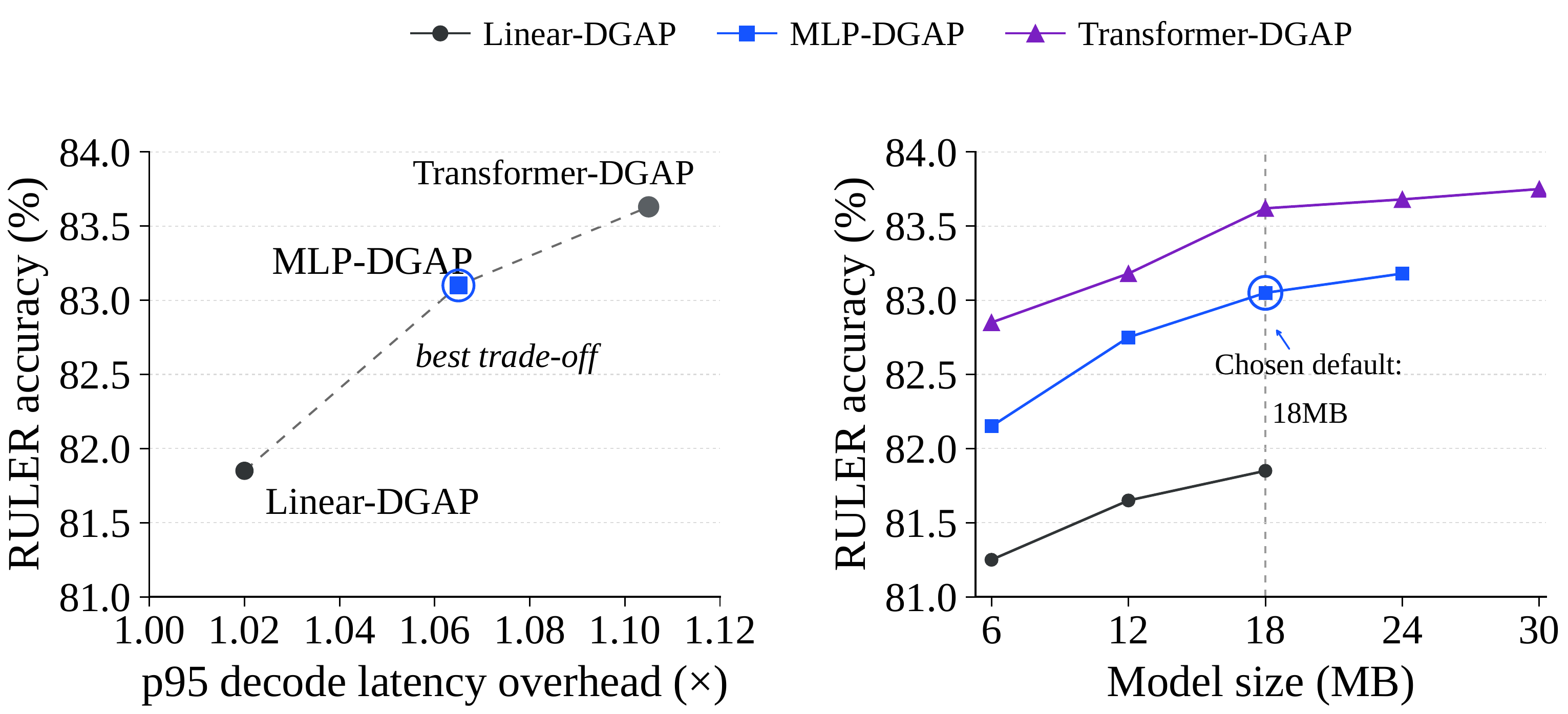}
    \caption{Accuracy--latency tradeoff of DGAP architectures. }
    \label{fig:architecture_tradeoff}
\end{figure}

\paragraph{Detector--corrector architecture.}
Fig.~\ref{fig:architecture_tradeoff} compares Linear, MLP, and Transformer DGAP variants in terms of accuracy, latency, and model size. Linear-DGAP is efficient but underfits the distribution-restoration objective, while Transformer-DGAP gives slightly higher accuracy at higher latency and larger size. MLP-DGAP provides the best accuracy--efficiency trade-off and is used as the default 18MB architecture. 

\begin{figure}[t]
    \centering
    \includegraphics[width=\linewidth,height=0.38\textheight,keepaspectratio]{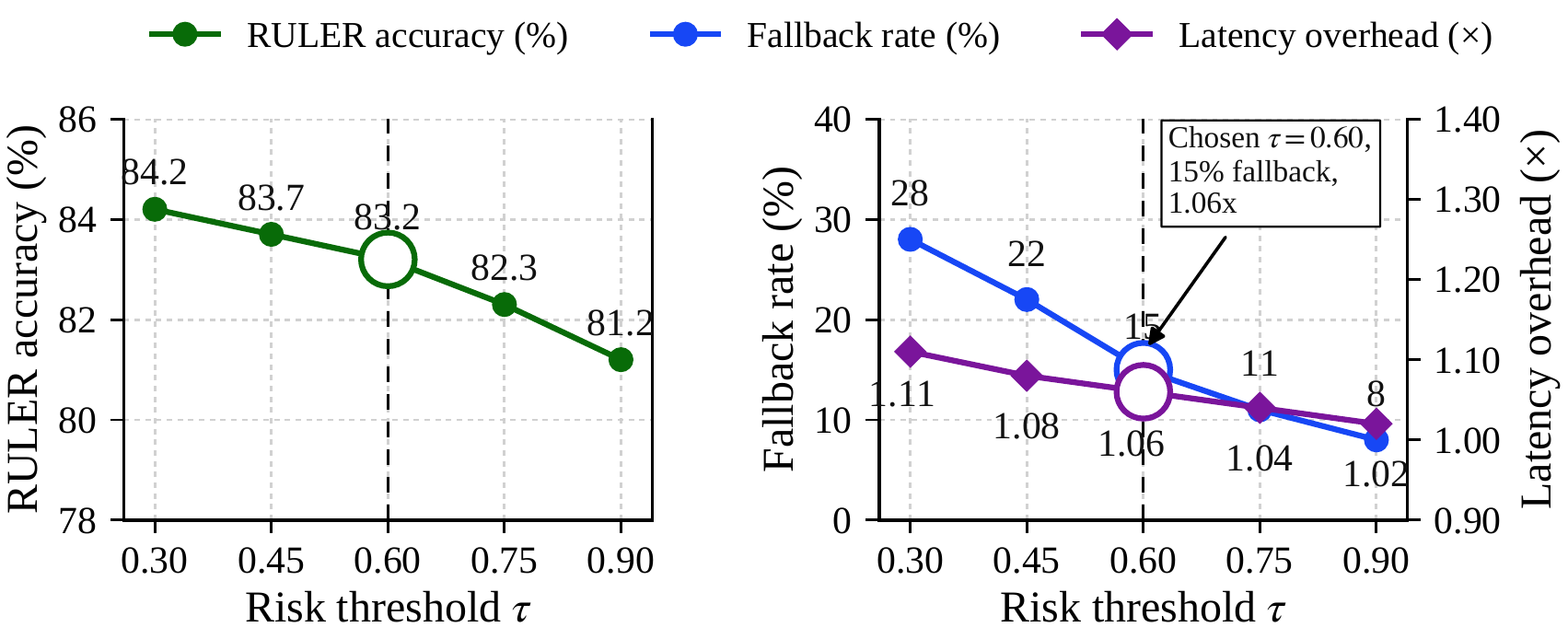}
    \caption{
    Effect of risk threshold \(\tau\) on K1V1+DGAP.
    }
    \label{fig:accuracy_threshold}
\end{figure}

\paragraph{Effect of risk threshold $\tau$.}
Finally, we study the impact of the trigger threshold $\tau$. As shown in Fig.~\ref{fig:accuracy_threshold}, lower $\tau$ increases correction frequency and improves restoration quality, but introduces higher latency due to more frequent activation. Higher $\tau$ reduces overhead but leaves more local distribution errors uncorrected. We set $\tau=0.60$ as the best trade-off, achieving strong accuracy with minimal runtime overhead.

\section{Conclusion}
\label{sec:conclusion}

We presented DGAP, a lightweight decode-time restoration layer for low-bit KV-cache quantized LLMs. Motivated by the observation that key quantization induces local top-$K$ candidate distribution reordering, DGAP detects recoverable distribution-risky steps and selectively restores the affected candidate region. Across benchmarks, DGAP recovers much of the performance loss under K1V1 while preserving the persistent low-bit KV-cache footprint and introducing only modest decode-time overhead. These results highlight the effectiveness of distribution-aware local correction as an alternative to increasing KV precision.

\section{Limitations and Future Work}
\label{sec:limitations}

DGAP focuses on algorithmic distribution restoration at decode time, while leaving system-level serving optimizations as future work. Its current implementation uses offline-calibrated thresholds and recovery windows derived from paired FP16 and quantized traces. Future extensions may explore online calibration and adaptive window selection for deployment-specific workloads, as well as kernel-level optimizations (e.g., Triton/CUDA fusion) to further reduce runtime overhead in large-scale serving systems.

\clearpage
\bibliography{aaai2027}
\end{document}